# A Comparison of 1-D and 2-D Deep Convolutional Neural Networks in ECG Classification

Yunan Wu, Feng Yang, Ying Liu, Xuefan Zha, Shaofeng Yuan

*Abstract*— Effective detection of arrhythmia is an important task in the remote monitoring of electrocardiogram (ECG). The traditional ECG recognition depends on the judgment of the clinicians' experience, but the results suffer from the probability of human error due to the fatigue. To solve this problem, an ECG signal classification method based on the images is presented to classify ECG signals into normal and abnormal beats by using two-dimensional convolutional neural networks (2D-CNNs). First, we compare the accuracy and robustness between one-dimensional ECG signal input method and two-dimensional image input method in AlexNet network. Then, in order to alleviate the overfitting problem in two-dimensional network, we initialize AlexNet-like network with weights trained on ImageNet, to fit the training ECG images and fine-tune the model, and to further improve the accuracy and robustness of ECG classification. The performance evaluated on the MIT-BIH arrhythmia database demonstrates that the proposed method can achieve the accuracy of 98% and maintain high accuracy within SNR range from 20 dB to 35 dB. The experiment shows that the 2D-CNNs initialized with AlexNet weights performs better than one-dimensional signal method without a large-scale dataset.

## I. INTRODUCTION

Nowadays, cardiac arrhythmia is a common disease with the bad consequences of sudden death or heart failure. ECG is basically the graph of voltage versus time which represents the electrical activity of myocardium during contraction and relaxation stages. The normal and abnormal classification of ECG signals plays an essential part in ECG monitoring and detection, such as wearable ECG recognition and monitor, as well as dynamic ECG diagnosis. Although the arrhythmias consist of different types, this paper is based on binary classification, whose methods applies to multi-classification as well. The traditional ECG recognition relies on the clinician's experience, but this method is limited by their time and space. Therefore, a better ECG classification method is necessary for clinicians to know the state of the patients.

*Research supported by the National Natural Science Foundation of China (NSFC：61771233，61271155).

Y. Wu is with the School of Biomedical Engineering, Southern Medical University, Guangdong, China (author phone: +86−13078140 057; e-mail: yunanwu.smu@gmail.com).

F. Yang is with the School of Biomedical Engineering, Southern Medical University, Guangdong, China (corresponding author phone: +86-13672401 148; e-mail: yangf@smu.edu.cn).

Y. Liu is with the School of Biomedical Engineering, Southern Medical University, Guangdong, China (e-mail: ying.liu.smu@gmail.com).

X. Zha is with the School of Biomedical Engineering, Southern Medical University, Guangdong, China (e-mail: xuefanzha.smu@gmail.com).

S. Yuan is with the School of Biomedical Engineering, Southern Medical University, Guangdong, China and Shanghai United Imaging Healthcare Co. Ltd., Shanghai, China (e-mail: shaofeng.yuan.smu@gmail.com).

In the past few years, many methods have been proposed to classify ECG signals automatically. For example, Pan and Tompkins [1] proposed the QRS detection algorithm based upon digital analyses of slope, amplitude, and width. In 2004, Shyu *et al.* [2] used wavelet transform (WT) to detect arrhythmia. However, these methods need to extract features manually, often missing much important information. Therefore, machine learning is introduced into the dectection of the ECG signals, such as Support Vector Machine (SVM) [3] and Artificial Neural Networks (ANNs) [4]. Ranaware and Deshpande [5] compared the detection performance of arrhythmia based on WT, ANNs and SVM and presented SVM of the best accuracy. In 2016, Kiranyaz *et al.* [6] proposed a novel way for patient-specific monitoring by using one-dimensional Convolutional Neural Network (1D-CNNs). Beyond this method, our paper proposed two-dimensional convolutional neural network (2D-CNNs) to make classification directly on the image.

In this papar, we input original ECG images into 2D-CNNs. With the weight initialization trained on the ImageNet, the accurancy and robustness performs better than 1D-CNNs. This method simulates the experienced clinicians in the way they judge the arrhythmia by simply processing the ECG image. To our knowledge, it's the first work where 2D-CNNs with transfer learning is used over ECG images. Fig. 1 shows the overall organization of this paper.

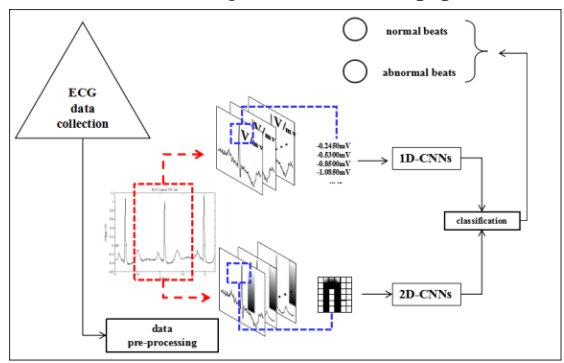

Figure 1. Overview of the proposed structure

## II. MATERIAL AND METHOD

### A. ECG Data Collection and Pre-processing

A well-known MIT-BIH arrhythmia database was adopted in our paper. It has 48 records in total, and each is in a ECG signal format over 30 minutes. The signal is annotated by cardiologist as types of normal and abnormal, in which includes the location of R wave peak and the type of the ECG beat. This paper uses the annotations to center R wave, so as to get a series of ECG heartbeat signals and images. Based on the whole section of ECG voltage, we predefine the upper and lower bounds of the coordinate axis, making sure each ECG beat is located at the right place, as shown in Fig. 2. In order to test the robustness of different methods, we add different gaussian noise to the signal. EMG interference is a kind of high-frequency noise, whose spectral characteristics are similar to the instantaneous white Gaussian noise [7], so we use white Gaussian noise within SNR range from 20 dB to 35 dB to disrupt the ECG signal, as shown in Fig. 2.

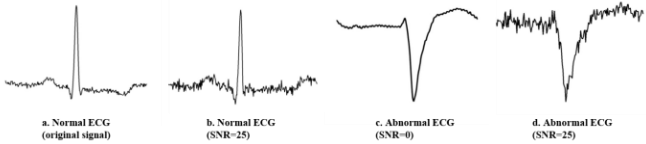

Figure 2. Pre-processed ECG data

### B. AlexNet-like 1D-CNNs and 2D-CNNS

The traditional neural network consists of three layers, including input layer, hidden layer and output layer. On the basis of the traditional neural network, the convolutional neural network (CNN) works as a feature extractor, adding up the convolution layer and the sub-sampling layer. It is the combination of the artificial neural network and back propagation algorithm, which simplifies the complexity of the model and reduces the parameters.

The introduction of AlexNet deepened the CNNs, making the training result more accurate. It has five convolution layers and three fully-connected layers, of which conv1, con2 and conv5 layers are connected with the max-pooling layers. Since fully-connected layers require the fixed dimensions of feature map, we set the input size 820 in length on 1D-CNNs and 256×256 on 2D-CNNs. Compared with common CNN, AlexNet has the following advantages: (i) The dropout layer is used after the fully-connected layer. Randomly ignoring some neurons in the training process can alleviate the over-fitting problem; (ii) Max-pooling layers are used to increase the richness of features; (iii) The nonlinear activation function ReLU is used to speed up the forward propagation process and solve the problem of gradient explosion. In this paper, we use the AlexNet-like CNNs model, whose structure is shown in Fig. 3.

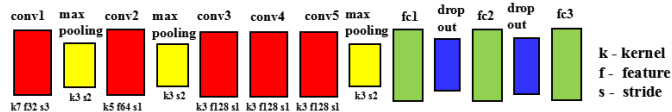

Figure 3. AlexNet-like structure

### C. Activation Functions

Activation function is working as the nonlinear factors to classify more complex data. In recent years, there are many popular activation functions, such as tanh, ReLU, ELU, SeLU, as the definitions in (1)~(4). Recently, Google Brain proposed a new self-gated activation function, Swish [8]. Swish (5) resembles ReLU since it has lower bound and no upper bound, but has the specific characteristics of smooth and nonmonotonicity, which is different from ReLU as shown in Fig. 4. In 1D-CNNs, by using the model structure similar to Fig. 3, we compare and analyze the accuracy of ECG classification with different activation functions, as shown in Fig. 4.

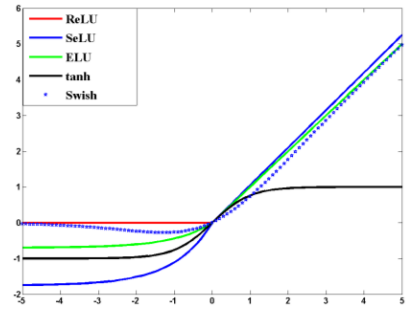

Figure 4. Different activation functions

$$\text{ReLU} = \max(0, x) \tag{1}$$

$$\text{ELU}(x) = \begin{cases} x & \text{if } x > 0 \\ \alpha(e^{-2x} - 1) & \text{if } x \leq 0 \end{cases} \tag{2}$$

$$\tanh(x) = \frac{1 - e^{-2x}}{1 + e^{-2x}} \tag{3}$$

$$\text{SeLU}(x) = \lambda \begin{cases} x & \text{if } x > 0 \\ \alpha(e^{-2x} - 1) & \text{if } x \leq 0 \end{cases} \tag{4}$$

$$\text{Swish}(x) = \frac{x}{1 + e^{-2x}} \tag{5}$$

### D. Transfer Learning and Fine Tuning

Traditional data mining and machine learning algorithms make predictions on the future data by using statistical models which are trained on previously collected labeled or unlabeled training data, but the premise is that the probability distributions of the training and testing data should be the same. In the training phase, weights in each convolutional layers of a CNN are initialized by values randomly sampled from a normal distribution with a zero mean and small standard deviation. For the ECG images, training a deep CNN from scratch is hard because (i) it requires a large amount of labeled training data and a great deal of expertise to ensure proper convergence; (ii) the process is time-consuming; (iii) it is likely to suffer from the problem of overfitting.

Transfer learning [9], allows the data used in training and testing to be different. Fine tuning [10] remains the choice for transfer learning with CNN: a model is pretrained on a source domain (where data is often abundant), the output layers of the model are adapted to the target domain, and the network is fine-tuned via back propagation. [11] have proposed that compared with random initialization of the model parameters, fine tuning improved accuracy and robustness of classification. In this paper, in order to alleviate the problem of overfitting in 2D-CNNs, the ECG images have been trained using a large labeled dataset from a different application, such as ImageNet.

ImageNet is a large visual database designed for the use in visual object recognition software research, which aims to populate the majority of the 80,000 synsets of WordNet with an average of 50 million cleanly labeled full resolution images. In this paper, we use the weights pretrained on ImageNet to fine-tune the model in order to take advantages of two-dimensional image input method in CNNs.

*E. Evaluation Metrics*

The performance of the proposed method is evaluated using the most popularly used metrics: sensitivity(Sen), specificity(Spe) and accuracy(Acc), which are defined in (6)~(8), where TP, TN, FP, and FN are true positive, true negative, false positive, and false negative, respectively. TP is the classification result where positive training data are evaluated as positive; TN is the classification result where negative training data are evaluated as negative; FP is the classification result where negative training data are evaluated as positive; and FN is the classification result where positive training data are evaluated as negative. Sensitivity measure correctly detected all real results, specificity measures the correctly rejected non-real results and accuracy is the measure of overall system performance.

$$\text{Sensitivity} = \frac{TP}{TP+FN} \times 100\% \quad (6)$$

$$\text{Specificity} = \frac{TN}{TN+FP} \times 100\% \quad (7)$$

$$\text{Accuracy} = \frac{TP+TN}{TP+TN+FN+FP} \times 100\% \quad (8)$$

III. EXPERIMENTAL RESULTS

In this paper, we use 1D-CNNs and 2D-CNNs to train the ECG signal pre-processed above. Two types of the input are set for each ECG beat: (1) a one-dimensional signal, the amplitude value of each ECG beat, is 820 (padding if not enough) in length; (2) a two-dimensional image, a 256 × 256 scaled pixel matrix, is the ECG waveform corresponding to the one-dimensional sequence. The training set is 5000 and testing set is 1500, and each beat occurs in either training or testing set. In the following experiment, we use five-fold cross validation and get the average results. The number of iterations is 2000. The training process adopts the stochastic gradient descent (SGD) which uses the variable learning rate strategy to keep the fast convergence of the model while restraining the problem of gradient vanishing. We have implemented the Alex-like network with Caffe. All experiments are performed on a computer with Intel I7-5930K CPU and a NVIDIA GTX1080 GPU.

*A. 1D-CNNs with Different Activation Functions*

The paper compares five activation functions in 1D-CNNs, tanh, ELU, SeLU, ReLU and Swish as shown in Table I. The classification shows a good performance on activation functions of Swish, ELU and ReLU, with the accuracy up to 96.00%, 95.50% and 95.40% respectively. The increase of noise signal interferes the accuracy. Swish, ReLU and tanh shows good robustness in different SNR, i.e. 1.3%,1.4% and 1.9% respectively. It is interesting that Swish is superior to other activation functions, which proves that not only Swish has better performance in large database and deep neural network [12], but also performs well in some of the small database, like ECG classification in this paper. Therefore, in 1D-CNNs, ECG signals can be best classified by the activation function of Swish, with both high accuracy and good robustness.

*B. 2D-CNNs with Fine Tuning*

With the decrease of SNR, the accuracy of testing set is declining, but the overall robustness is good. As shown in Table I., the average range of the accuracy is within 2.60%. However, in the network with random weight initialization, the accuracy of the 1D signal input is always higher than that of the 2D image input. The result is the same with robustness. Because the number of input parameters in 1D-CNNs is much less than that in 2D-CNNs, the model with more parameters and higher complexity is more likely to have overfitting problems due to the limited data, affecting the accuracy of the 2D-CNNs model. Therefore, we use the weights trained on ImageNet, to fit the ECG image and fine-tune to alleviate the problem of overfitting. Fig. 5 shows the accuracy vs number of iteration curve for testing set. In Table I., after using the weights initialed on ImageNet, the accuracy reaches up to 98.00%, and the robustness fluctuated at 1.25%, both exceeding the 1-D input method.

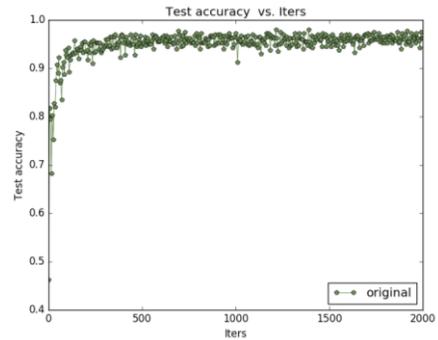

Figure 5. Testing accuracy vs number of iteration

*C. Evaluation Performance*

Table II. shows the classification results in comparison with the performance in 1D-CNNs (different activation functions) and 2D-CNNs (different weight initialization). In general, the classification with weight trained on ImageNet in 2D-CNNs is superior, i.e., 99.00%, 96.50% and 98.00%, for sensitivity, specificity and accuracy, respectively. Compared with random weight initialization, is 96.67%, 91.00% and 94.40%. In general, the Swish activation performs best in 1D-CNNs, i.e. 98.33%, 92.50% and 96.00%, which is still worse than those in 2D-CNNs with ImageNet weight to do fine tuning in 2D-CNNs. In Table II., for all types of the model, we can find the reason why the specificity performs worse than the sensitivity. Compared with normal beats, abnormal ECG signal types are more complex, including ventricular flutter，atrial fibrillation, ventricular escape beat and unclassifiable beat, which make it difficult for the model to classify. Table III. shows that the proposed method could reach the best accuracy among the previous state of the art.

TABLE I. ACCURACY OF DIFFERENT TYPES OF MODELS

| Type | Signal-Noise Ratio (dB) | | | | |
|---|---|---|---|---|---|
| | *none* | *35* | *30* | *25* | *20* |
| tanh-1D | 94.90 | 94.80 | 94.50 | 94.30 | 93.00 |
| ELU-1D | 95.50 | 95.10 | 94.90 | 93.70 | 92.90 |
| SeLU-1D | 94.40 | 93.80 | 93.50 | 92.00 | 91.60 |
| ReLU-1D | 95.40 | 95.30 | 94.80 | 94.50 | 94.00 |
| Swish-1D | **96.00** | **95.60** | **95.30** | **95.30** | **94.70** |
| Random-2D | 94.40 | 94.00 | 93.90 | 93.70 | 92.80 |
| ImageNet-2D | **98.00** | **97.50** | **97.00** | **96.75** | **96.75** |

TABLE II. EVALUATION RESULTS

| Metrics | Type | | | |
|---|---|---|---|---|
| | *Swish-1D* | *ReLU-1D* | *Random-2D* | *ImageNet-2D* |
| Sensitivity | 98.33 | 98.16 | 96.67 | **99.00** |
| Specificity | 92.50 | 91.50 | 91.00 | **96.50** |
| Accuracy | 96.00 | 95.40 | 94.40 | **98.00** |

TABLE III. COMPARISON OF ECG BEAT CLASSIFICATION METHODS ON MIT-BIH ARRHYTHMIA DATABASE (OPTIMAL ARE HIGHLIGHTED)

| Methods | Classifier | Class | Accuracy |
|---|---|---|---|
| Fujita *et al.* [12] | CNN | 2 | 94.90% |
| Inan *et al.* [13] | NN | 2 | 95.20% |
| Swish-1D | 1D-CNN | 2 | 96.00% |
| **Our method** | **2D-CNN** | **2** | **98.00%** |

IV. CONCLUSION

In this work, we compare the classification performance between one-dimensional signal input and two-dimensional image input in convolutional neural networks. In 1D-CNNs, the activation of Swish has higher accuracy and robustness compared with other activation functions, which can be used in ECG classification. In the alike 2D-CNNs model, after finding that the random initialization performs worse than 1D-CNNs, we use the weight pretrained on the ImageNet to initialize the model, effectively alleviating the problem of overfitting. The accuracy of this proposed method can reach up to 98.00%. Compared with traditional classification methods, Image input method in 2D-CNNs doesn't need to extract features manually. Compared with signal input method in 1D-CNNs, 2D-CNNs can do fine tuning with large database, achieving higher accuracy and robustness. The comparison with state-of-art methods demonstrates that the proposed methods achieve the best performance.

Moreover, the proposed method applies well on another five-class classification and get the accuracy of 94.5%. In the future work, we can try to input a section of ECG signals instead of the ECG beats, so as to achieve the automatic and real-time classification.

ACKNOWLEDGMENT

The research is supported by the National Natural Science Foundation of China (NSFC：61771233，61271155). The authors also wish to acknowledge the faculties in the department of Electronic Technology, School of Biomedical Engineering, Southern Medical University, for their support and assistance. The research work is carried out on ECG signals loaded from MIT-BIH database, website: www.ecg.mit-bih.edu.